\documentclass{llncs}
\usepackage{graphicx}

\usepackage{graphicx}
\usepackage{soul}
\graphicspath{ {./figs/} }

\usepackage{tikz}
\usetikzlibrary{shapes,decorations,arrows}

\makeatletter
\newcommand{\printfnsymbol}[1]{%
  \textsuperscript{\@fnsymbol{#1}}%
}
\makeatother

\begin{document}
\title{Transformer Assisted Convolutional Network for Cell Instance Segmentation}
%
%
\author{Deepanshu Pandey\thanks{Authors contributed equally} \and
Pradyumna Gupta\printfnsymbol{1} \and
Sumit Bhattacharya\printfnsymbol{1} \and
Aman Sinha \and
Rohit Agarwal}


%
%
\institute{Indian Institute of Technology (Indian School of Mines), Dhanbad, Jharkhand, India }


%
\maketitle              

\begin{abstract}
Region proposal based methods like R-CNN and Faster R-CNN models have proven to be extremely successful in object detection and segmentation tasks.  Recently, Transformers have also gained popularity in the domain of Computer Vision, and are being utilised to improve the performance of conventional models.
In this paper, we present a relatively new transformer based approach to enhance the performance of the conventional convolutional feature extractor in the existing region proposal based methods. Our approach merges the convolutional feature maps with transformer-based token embeddings by applying a projection operation similar to self-attention in transformers. The results of our experiments show that transformer assisted feature extractor achieves a significant improvement in mIoU (mean Intersection over Union) scores compared to vanilla convolutional backbone.



\end{abstract}

\section{Introduction} \label{1}
 Segmentation of cells in microscopic images is an important task in the fields of biological research, clinical practice and disease diagnosis.  Robust plasma cell segmentation is the first step\cite{10.1371/journal.pone.0207908} towards detection of cancerous cells in case of Multiple Myeloma(MM), a form of blood cancer. Given the voluminous data available, there is a growing need for automated algorithms and tools for the analysis of these cells. In addition to that, owing to varying intra-cellular and inter-cellular dynamics and structural traits of cells, there is a persistent need for more accurate and efficient segmentation models.

Cell segmentation is challenging due to many factors like unclear background, cell division, distorted cell shapes and clustered cells. Convolutional Neural Networks(CNNs) have been deployed, at large, for biomedical image segmentation and have shown good results despite the fact that they fail to capture relative spatial information of cells in a slide image. An architecture that is still under-explored in the domain of biomedical image segmentation is that of the Transformer\cite{vaswani2017attention}. An advantage of Transformer is that the spatial information of features is preserved when compared with CNNs. For the purpose of this paper, we adapt the implementation of Visual Transformers as described by \cite{wu2020visual} in our experiments, which is explained later.

To improve the performance of CNN models, we propose a method of fusing ViT\cite{dosovitskiy2020image} with CNN backbones for instance segmentation of plasma cells of patients diagnosed with MM. This fusion strategy produce masks that are comparatively more robust than the ones obtained using only CNNs.

\section{Dataset} \label{2}
For our experiments, we used the SegPC-2021\cite{7np1-2q42-21} dataset which consists of microscopic images that were captured\cite{Gupta2018} from bone marrow aspirate slides of patients diagnosed with MM, and cell instance masks for each image corresponding to the malignant cells present in the image. The slides for capturing the images were stained with the help of Jenner-Giemsa stain using the stain color normalization pipeline described in \cite{GUPTA2020101788}. The dataset comprises of images in two shapes, \( 2040 \times 1536 \) and \(1920 \times 2560\), depending on the hardware used to take these images. The overall dataset is divided into three sets : training set (298 images), validation set (200 images) and test set (277 images).

\begin{figure}
\includegraphics[height = 21mm, width=\textwidth]{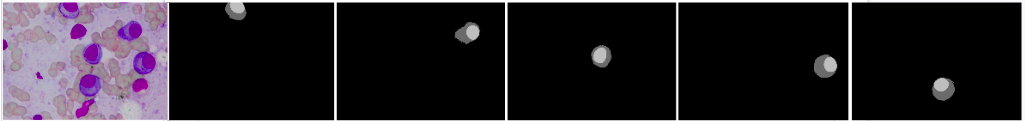}
\caption{Slide image and corresponding Ground Truth masks.} \label{fig1}
\end{figure}

\section{Related Works} \label{3}
Pixel-level analysis of images has been greatly accelerated by the recent advances in deep learning as the research continues to move from coarser towards finer inference\cite{Hafiz_2020}.  Segmentation of objects in images by pixel level classification is one such application which has been extensively applied in various biomedical and business problems. Segmenting cells from microscope images using deep learning is being explored by a lot of studies lately, including semantic segmentation of cells\cite{DBLP:journals/corr/RonnebergerFB15,8644754} which associates every pixel of an image with a class label as compared to instance level segmenting of cells\cite{DBLP:journals/corr/abs-1805-00500,10.1007/978-3-030-11024-6_29,8363596} which treats different cells as distinct instances. While our approach employs models specific to instance segmentation tasks, there have been studies which extend semantic segmentation models using various training and post-processing strategies for instance separation including, instance aware embeddings\cite{DBLP:journals/corr/abs-1711-09060,DBLP:journals/corr/abs-1708-02551,DBLP:journals/corr/FathiWRWSGM17} and instance separation techniques\cite{article,DBLP:journals/corr/BaiU16,nishimura2019weakly}.

Region proposal based methods, namely the R-CNN family of models\cite{girshick2014rich,girshick2015fast,ren2016faster}, have been providing significant accuracy gains on various localization tasks ever since their release, as compared to other architectures. The addition of mask prediction branch to the Feature Proposal Network(FPN) of Faster RCNN by \cite{he2018mask} led to the Mask RCNN architecture which has been used by various studies like \cite{DBLP:journals/corr/abs-1805-00500} and \cite{10.1007/978-3-030-69756-3_5} for cell instance segmentation. Due to the success of the RCNN models, most of the state of the art object detection and segmentation approaches released later, including the ones used in this paper, Cascade Mask RCNN\cite{cai2019cascade} and DetectoRS\cite{qiao2020detectors}, were inspired from these RCNN models and share a similar architecture. The architecture can be broadly divided into 3 components : Feature Pyramid Network(FPN, used as the CNN backbone), Region Proposal Network(RPN) and Region of Interest(RoI) heads.

Apart from the convolution based models, the Transformers\cite{vaswani2017attention} have been influencing the landscape of deep learning ever since their release. Originally released for Natural Language Processing (NLP) tasks, recent studies like \cite{dosovitskiy2020image}, \cite{esser2021taming} and \cite{carion2020endtoend} have shown great results in various vision related tasks also. Among these, the study by \cite{wu2020visual} on Visual Transformer, where the authors use a transformer to process the token-based representations obtained from the feature maps of the CNN encoder and later project these tokens to obtain the mask predictions, have greatly inspired our methods, which is described later.

\section{Methodology} \label{4}
\subsection{Cascade Mask RCNN}
Originally introduced in \cite{cai2019cascade}, the cascade R-CNN architecture is a multi-level extension of the R-CNN\cite{girshick2014rich} model, which uses the idea of region proposals for localisation of objects in an image. 

Cascade R-CNN uses a number of subsequent network heads, which are generally sequential in nature, deriving information and improving the results of the previous head. Cascade Mask R-CNN adds an extra mask head to the model to utilise it for instance segmentation tasks. The model architecture can be divided into three parts - a backbone feature extractor, an RPN and ROI Heads (or Box Heads). The backbone network takes the input image and provides feature maps which are used by the RPN to propose potential object regions in the image. The outputs of the RPN along with the feature maps are then fed to the ROI heads which crop the feature maps using the RPN proposal boxes to obtain fine-tuned object bounding boxes, object classes and segmentation masks.

\subsection{DetectoRS}
The recently released DetectoRS model by \cite{qiao2020detectors} has shown state of the art results on various datasets including the COCO dataset\cite{lin2015microsoft}, owing to the two architectural contributions : Recursive Feature Pyramids and Switchable Atrous Convolutions. Recursive Feature Pyramids apply the idea of thinking and looking twice at the backbone level of the network by stacking the FPN blocks horizontally multiple times. The inputs of the intermediate blocks is based on the feedback received from the outputs of previous blocks, before finally taking the outputs of the final block and propagating them to other parts of the network. Switchable Atrous Convolution makes use of a learn-able switch (or gate) to switch between different atrous (or dilation) rates which, thus, helps in dynamically changing the receptive field of an otherwise standard convolution operation. The authors finally apply these ideas to the Hybrid Task Cascade\cite{chen2019hybrid} model to obtain DetectoRS.

\subsection{Transformer Assisted Convolutions}

Our proposed methodology revolves around the idea of enhancing the performance of convolutional feature extractors by merging them with the token-based representations generated by the transformer models. The idea, inspired from \cite{wu2020visual}, is to use both convolutional and transformer based networks (both equal in depth) to first create convolutional and transformer-based feature maps independently for a given image. The corresponding feature maps at each level are then combined using a projection layer similar to the one defined in \cite{wu2020visual}. Treating this as a self-attention mechanism, the projections of these convolutional feature maps and the transformer tokens act as query and key respectively, the product of which determines how the information in the transformer tokens is to be projected (which are to be treated as value). More formally, we define- 

\[S_i := C_i + {softmax}_L((C_{i}W_{q})(T_{i}W_{k}))T_{i} \]

where \(T_i  \in  R^{L \times E}\) is the \(i^{th}\) transformer encoder state, \(C_i  \in  R^{HW \times E}\) is the \(i^{th}\) convolutional encoder output (or the \(i^{th}\) convolutional feature map) and \(S_i  \in  R^{HW \times E}\) is the \(i^{th}\) final merged state. Here, L is the latent dimension for the transformer defining the number of transformer tokens to be used. E is the number of channels in the output feature maps, hereby referred to as the embedding dimension. H and W are the height and width of the feature map under consideration. \(W_q  \in  R^{E \times E}\) and \(W_k  \in  R^{E \times E}\) are learnable weights used for projection.

Unlike the Visual Transformer which uses only the final feature map for projection of transformer tokens to segmentation maps, our approach tries to combine the convolutional features maps at different depths with their corresponding transformer tokens at the same depth. This is done with the idea that with increasing depth, feature extractors learn increasingly more complex patterns and it would be more meaningful to merge a deeper convolutional feature map with a transformer token at similar depth. Also, \cite{wu2020visual} uses convolutional features maps to generate the initial transformer token and uses a recurrent operation to generate subsequent tokens. However, we simply use an independent ViT module to capture the pixel-wise information in transformer tokens which is later merged with its corresponding convolutional feature map in the projection layer.

A generalised architecture for our proposed model backbone, the transformer-assisted  feature extractor, is given in Fig2.




\begin{figure}[ht]
\includegraphics[height= 78mm, width=123mm]{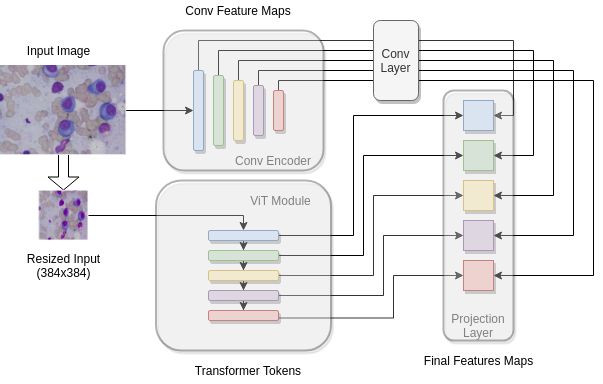}
\caption{Transformer-assisted convolutional feature extractor}
\label{fig}
\end{figure}

\subsection{Models and Architecture}

The ViT, pretrained on ImageNet\cite{5206848}, was used to extract transformer tokens from input images. The input images were resized to match the input dimensions accepted by ViT. This transformer model was merged with two types of instance segmentation models, namely Cascade Mask R-CNN and DetectoRS, both of which were pretrained on COCO Instance Segmentation dataset. The convolutional feature extractors (encoders) in these models were also replaced with EfficientNet-b5 \footnote{publically available in segmentation-models-pytorch \cite{Yakubovskiy2019} library as timm-efficientnet-b5} \cite{tan2020efficientnet} backbones pretrained on ImageNet. 

\section{Results} \label{5}

\subsection{Experimental Setup}

The models were trained on NVIDIA Tesla T4 and P100 GPUs provided by Google Colaboratory. We make our code publicly 
\footnote{ 
\url{https://github.com/dsciitism/SegPC-2021}}
available. 

\begin{table}
\centering
\caption{Results obtained using various models on the test set. 
Here, DRS refers to DetectoRS , CMRCNN-X152 refers to Cascade Mask RCNN 152 , Effb5 refers to EfficientNet b5 as backbone and trans refers to the variant of the corresponding model with EfficientNet b5 and Transformer as backbones.
}\label{tab1}
\begin{tabular}{| l | l |}
\hline 
\textbf{Model}
& \textbf{mIoU Score} \\

\hline

DRS  &  0.9219 \\
CMRCNN-X152 &  0.9179 \\

\hline

DRS with Effb5 &  0.8793 \\
CMRCNN-X152 with Effb5 &  0.9038 \\

\hline

DRS with Effb5 and Transformer &  \textbf{0.9273} \\
CMRCNN-X152 with Effb5 and Transfomer &  \textbf{0.9281} \\
\hline


\hline

\end{tabular}
\end{table}

The Cascade Mask RCNN model\footnote{Publicly available as cascade\_mask\_rcnn\_X\_152\_32x8d\_FPN\_IN5k\_gn\_dconv in the detectron2 library\cite{wu2019detectron2} } used in our experiments comprises of an FPN based ResNeXt-152 \(32 \times 8d\) \cite{xie2017aggregated} backbone pretrained on Imagenet-5k. The experiments were done with batch size of 1 due to computational constraints. The initial learning rate was set to be 1e-3 with Cosine Annealing Scheduler. Each of the models was trained for 20000 iterations with 3000 warm-up iteration. Augmentations used for training are random flip, rotation and random resize(chosen from a set of predefined sizes).

For DetectoRS\footnote{Publicly available as DetectoRS in the mmdetection library\cite{mmdetection}}, we used the original implementation as described by \cite{qiao2020detectors} pretrained on the COCO dataset. Due to computational constraints, the model was trained with a batch size of 1 using a step decay learning rate scheduler with a base learning rate of 2.5e-4. The model was trained for 10000 iterations using SGD optimizer with random flip and random resize (chosen from a set of predefined sizes) as image augmentations.

The probability threshold for ROI heads in the models was set to be 0.001.
 The encoder depth was kept 5 for all the models. It is to be noted that using EfficientNet-b5 backbone \footnote{ Despite the superior performance of default convolutional backbone over EfficientNet-b5 in both CMRCNN and DRS, it was not possible for us to train these along with our proposed transformer model due to their large size and limited computational resources available to us. Therefore, we trained our transformer-assisted models with much lighter EfficientNet-b5 as backbone} in the convolutional feature extractor gives 5 outputs with channel dimensions \{3, 48, 40, 64, 176, 512\}, whereas the projection layer defined in section 4.1 requires the channel dimension to be consistent for all the encoder outputs and equal to the embedding dimension defined for the transformer. An intermediate convolution operation is therefore applied on each of the outputs of the convolutional feature extractor in order to make the channels equal to the required embedding dimension.











\subsection{Model Performance}
The models were evaluated on the test set with mIoU(mean Intersection over Union) as the metric, and their performance has been tabulated in Table \ref{tab1}.We report the scores of the normal and transformer assisted variants of DetectoRS and Cascade Mask RCNN with EfficientNet-b5 backbone, along with the scores of the default backbones. As has been explained earlier, due to the computational constraints, we are not able to report the scores of the transformer assisted variants of the default convolutional backbones of DetectoRS and Cascade Mask RCNN. However, we strongly believe that the transformer assisted default backbones of these models would have given superior performance. The difference between the scores of the normal and transformer assisted version of the EfficienNet-b5 backbone is an evidence to support our claim. Furthermore, the scores of the transformer assisted version of the much lighter EfficientNet-b5 backbone is already an improvement over the original architectures.












\section{Conclusion} \label{6}
Our results show that the transformer-assisted convolutional backbones outperform the vanilla convolutional backbones and result in finer segmentation maps. Also, the approach is not limited to any individual transformer model, but can be generalised to any transformer based encoder, by treating the projection layer as a self attention operation between the convolutional feature maps and the transformer tokens. This approach of transformer assisted convolutional backbones can also be used for generating improved feature maps for other computer vision tasks like classification, segmentation, image captioning etc.

\bibliographystyle{unsrt}
\bibliography{All/biblio}

\begin{thebibliography}{10}

\bibitem{10.1371/journal.pone.0207908}
Anubha Gupta, Pramit Mallick, Ojaswa Sharma, Ritu Gupta, and Rahul Duggal.
\newblock Pcseg: Color model driven probabilistic multiphase level set based
  tool for plasma cell segmentation in multiple myeloma.
\newblock {\em PLOS ONE}, 13(12):1--22, 12 2018.

\bibitem{vaswani2017attention}
Ashish Vaswani, Noam Shazeer, Niki Parmar, Jakob Uszkoreit, Llion Jones,
  Aidan~N. Gomez, Lukasz Kaiser, and Illia Polosukhin.
\newblock Attention is all you need, 2017.

\bibitem{wu2020visual}
Bichen Wu, Chenfeng Xu, Xiaoliang Dai, Alvin Wan, Peizhao Zhang, Zhicheng Yan,
  Masayoshi Tomizuka, Joseph Gonzalez, Kurt Keutzer, and Peter Vajda.
\newblock Visual transformers: Token-based image representation and processing
  for computer vision, 2020.

\bibitem{dosovitskiy2020image}
Alexey Dosovitskiy, Lucas Beyer, Alexander Kolesnikov, Dirk Weissenborn,
  Xiaohua Zhai, Thomas Unterthiner, Mostafa Dehghani, Matthias Minderer, Georg
  Heigold, Sylvain Gelly, Jakob Uszkoreit, and Neil Houlsby.
\newblock An image is worth 16x16 words: Transformers for image recognition at
  scale, 2020.

\bibitem{7np1-2q42-21}
Anubha Gupta; Ritu Gupta; Shiv Gehlot;~Shubham Goswami.
\newblock Segpc-2021: Segmentation of multiple myeloma plasma cells in
  microscopic images, 2021.

\bibitem{Gupta2018}
Anubha Gupta, Pramit Mallick, Ojaswa Sharma, Ritu Gupta, and Rahul Duggal.
\newblock Pcseg: Color model driven probabilistic multiphase level set based
  tool for plasma cell segmentation in multiple myeloma.
\newblock {\em PLoS ONE}, 13(12):1--22, 2018.

\bibitem{GUPTA2020101788}
Anubha Gupta, Rahul Duggal, Shiv Gehlot, Ritu Gupta, Anvit Mangal, Lalit Kumar,
  Nisarg Thakkar, and Devprakash Satpathy.
\newblock Gcti-sn: Geometry-inspired chemical and tissue invariant stain
  normalization of microscopic medical images.
\newblock {\em Medical Image Analysis}, 65:101788, 2020.

\bibitem{Hafiz_2020}
Abdul~Mueed Hafiz and Ghulam~Mohiuddin Bhat.
\newblock A survey on instance segmentation: state of the art.
\newblock {\em International Journal of Multimedia Information Retrieval},
  9(3):171–189, Jul 2020.

\bibitem{DBLP:journals/corr/RonnebergerFB15}
Olaf Ronneberger, Philipp Fischer, and Thomas Brox.
\newblock U-net: Convolutional networks for biomedical image segmentation.
\newblock {\em CoRR}, abs/1505.04597, 2015.

\bibitem{8644754}
T.~{Tran}, O.~{Kwon}, K.~{Kwon}, S.~{Lee}, and K.~{Kang}.
\newblock Blood cell images segmentation using deep learning semantic
  segmentation.
\newblock In {\em 2018 IEEE International Conference on Electronics and
  Communication Engineering (ICECE)}, pages 13--16, 2018.

\bibitem{DBLP:journals/corr/abs-1805-00500}
Jeremiah~W. Johnson.
\newblock Adapting mask-rcnn for automatic nucleus segmentation.
\newblock {\em CoRR}, abs/1805.00500, 2018.

\bibitem{10.1007/978-3-030-11024-6_29}
Jingru Yi, Pengxiang Wu, Menglin Jiang, Daniel~J. Hoeppner, and Dimitris~N.
  Metaxas.
\newblock Instance segmentation of neural cells.
\newblock In Laura Leal-Taix{\'e} and Stefan Roth, editors, {\em Computer
  Vision -- ECCV 2018 Workshops}, pages 395--402, Cham, 2019. Springer
  International Publishing.

\bibitem{8363596}
Jingru Yi, Pengxiang Wu, Daniel~J. Hoeppner, and Dimitris Metaxas.
\newblock Pixel-wise neural cell instance segmentation.
\newblock In {\em 2018 IEEE 15th International Symposium on Biomedical Imaging
  (ISBI 2018)}, pages 373--377, 2018.

\bibitem{DBLP:journals/corr/abs-1711-09060}
Thomio Watanabe and Denis~F. Wolf.
\newblock Distance to center of mass encoding for instance segmentation.
\newblock {\em CoRR}, abs/1711.09060, 2017.

\bibitem{DBLP:journals/corr/abs-1708-02551}
Bert~De Brabandere, Davy Neven, and Luc~Van Gool.
\newblock Semantic instance segmentation with a discriminative loss function.
\newblock {\em CoRR}, abs/1708.02551, 2017.

\bibitem{DBLP:journals/corr/FathiWRWSGM17}
Alireza Fathi, Zbigniew Wojna, Vivek Rathod, Peng Wang, Hyun~Oh Song, Sergio
  Guadarrama, and Kevin~P. Murphy.
\newblock Semantic instance segmentation via deep metric learning.
\newblock {\em CoRR}, abs/1703.10277, 2017.

\bibitem{article}
Jos Roerdink and A.~Meijster.
\newblock The watershed transform: Definitions, algorithms and parallelization
  strategies.
\newblock {\em Fundam Inf}, 41, 10 2003.

\bibitem{DBLP:journals/corr/BaiU16}
Min Bai and Raquel Urtasun.
\newblock Deep watershed transform for instance segmentation.
\newblock {\em CoRR}, abs/1611.08303, 2016.

\bibitem{nishimura2019weakly}
Kazuya Nishimura, Dai Fei~Elmer Ker, and Ryoma Bise.
\newblock Weakly supervised cell instance segmentation by propagating from
  detection response, 2019.

\bibitem{girshick2014rich}
Ross Girshick, Jeff Donahue, Trevor Darrell, and Jitendra Malik.
\newblock Rich feature hierarchies for accurate object detection and semantic
  segmentation, 2014.

\bibitem{girshick2015fast}
Ross Girshick.
\newblock Fast r-cnn, 2015.

\bibitem{ren2016faster}
Shaoqing Ren, Kaiming He, Ross Girshick, and Jian Sun.
\newblock Faster r-cnn: Towards real-time object detection with region proposal
  networks, 2016.

\bibitem{he2018mask}
Kaiming He, Georgia Gkioxari, Piotr Dollár, and Ross Girshick.
\newblock Mask r-cnn, 2018.

\bibitem{10.1007/978-3-030-69756-3_5}
Seiya Fujita and Xian-Hua Han.
\newblock Cell detection and segmentation in microscopy images with improved
  mask r-cnn.
\newblock In Imari Sato and Bohyung Han, editors, {\em Computer Vision -- ACCV
  2020 Workshops}, pages 58--70, Cham, 2021. Springer International Publishing.

\bibitem{cai2019cascade}
Zhaowei Cai and Nuno Vasconcelos.
\newblock Cascade r-cnn: High quality object detection and instance
  segmentation, 2019.

\bibitem{qiao2020detectors}
Siyuan Qiao, Liang-Chieh Chen, and Alan Yuille.
\newblock Detectors: Detecting objects with recursive feature pyramid and
  switchable atrous convolution, 2020.

\bibitem{esser2021taming}
Patrick Esser, Robin Rombach, and Björn Ommer.
\newblock Taming transformers for high-resolution image synthesis, 2021.

\bibitem{carion2020endtoend}
Nicolas Carion, Francisco Massa, Gabriel Synnaeve, Nicolas Usunier, Alexander
  Kirillov, and Sergey Zagoruyko.
\newblock End-to-end object detection with transformers, 2020.

\bibitem{lin2015microsoft}
Tsung-Yi Lin, Michael Maire, Serge Belongie, Lubomir Bourdev, Ross Girshick,
  James Hays, Pietro Perona, Deva Ramanan, C.~Lawrence Zitnick, and Piotr
  Dollár.
\newblock Microsoft coco: Common objects in context, 2015.

\bibitem{chen2019hybrid}
Kai Chen, Jiangmiao Pang, Jiaqi Wang, Yu~Xiong, Xiaoxiao Li, Shuyang Sun,
  Wansen Feng, Ziwei Liu, Jianping Shi, Wanli Ouyang, Chen~Change Loy, and
  Dahua Lin.
\newblock Hybrid task cascade for instance segmentation, 2019.

\bibitem{5206848}
J.~{Deng}, W.~{Dong}, R.~{Socher}, L.~{Li}, {Kai Li}, and {Li Fei-Fei}.
\newblock Imagenet: A large-scale hierarchical image database.
\newblock In {\em 2009 IEEE Conference on Computer Vision and Pattern
  Recognition}, pages 248--255, 2009.

\bibitem{Yakubovskiy2019}
Pavel Yakubovskiy.
\newblock Segmentation models pytorch.
\newblock \url{https://github.com/qubvel/segmentation\_models.pytorch}, 2020.

\bibitem{tan2020efficientnet}
Mingxing Tan and Quoc~V. Le.
\newblock Efficientnet: Rethinking model scaling for convolutional neural
  networks, 2020.

\bibitem{wu2019detectron2}
Yuxin Wu, Alexander Kirillov, Francisco Massa, Wan-Yen Lo, and Ross Girshick.
\newblock Detectron2.
\newblock \url{https://github.com/facebookresearch/detectron2}, 2019.

\bibitem{xie2017aggregated}
Saining Xie, Ross Girshick, Piotr Dollár, Zhuowen Tu, and Kaiming He.
\newblock Aggregated residual transformations for deep neural networks, 2017.

\bibitem{mmdetection}
Kai Chen, Jiaqi Wang, Jiangmiao Pang, Yuhang Cao, Yu~Xiong, Xiaoxiao Li,
  Shuyang Sun, Wansen Feng, Ziwei Liu, Jiarui Xu, Zheng Zhang, Dazhi Cheng,
  Chenchen Zhu, Tianheng Cheng, Qijie Zhao, Buyu Li, Xin Lu, Rui Zhu, Yue Wu,
  Jifeng Dai, Jingdong Wang, Jianping Shi, Wanli Ouyang, Chen~Change Loy, and
  Dahua Lin.
\newblock {MMDetection}: Open mmlab detection toolbox and benchmark.
\newblock {\em arXiv preprint arXiv:1906.07155}, 2019.

\end{thebibliography}

\end{document}